\documentclass[conference]{IEEEtran}
\usepackage[dvips]{graphicx} 
\usepackage{epsfig}  
\usepackage{verbatim}
\usepackage{cite}
\usepackage{amsmath}
\usepackage{algorithmic}
\usepackage{mathtools, cuted}
\usepackage{lipsum, color}

\usepackage{array}
\usepackage{caption}
\usepackage{subcaption}
\ifCLASSOPTIONcompsoc  \usepackage[caption=false,font=normalsize,labelfont=sf,textfont=sf]{subfig}
\else  \usepackage[caption=false,font=footnotesize]{subfig}
 \fi
\begin{document}

\title{Color Aesthetics: Fuzzy based User-driven Method for Harmony and Preference Prediction  
\\ }


\author{\IEEEauthorblockN{Pakizar Shamoi}
\IEEEauthorblockA{Department of Information Systems Management,\\ Kazakh-British Technical University\\
Almaty, Kazakhstan\\
Email: pakita.shamoi@gmail.com}
\and
\IEEEauthorblockN{Atsushi Inoue}
\IEEEauthorblockA{Department of Computer Science,\\ Eastern Washington University\\
Washington, USA\\
Email: inoueatsushij@gmail.com}
\and
\IEEEauthorblockN{Hiroharu Kawanaka}
\IEEEauthorblockA{Graduate School of Engineering, \\Mie University\\
Tsu, Japan\\
Email: kawanaka@elec.mie-u.ac.jp}

}

\maketitle

\begin{abstract}
Color is the most important intrinsic sensory feature that has a powerful impact on product sales. Color is even responsible for raising the aesthetic senses in our brains.  Account for individual differences is  crucial in color aesthetics. It requires user-driven mechanisms for various e-commerce applications.
We propose a method for quantitative evaluation of all types of perceptual responses to color(s): distinct color preference, color harmony, and color combination preference . Preference for color schemes can be predicted combining preferences for the basic colors and ratings of color harmony. Harmonious pallets are extracted from big data set using comparison algorithms based on fuzzy similarity and grouping.The proposed model results in useful predictions of harmony and preference of multicolored images. For example, in the context of apparel coordination, it allows predicting a preference for a look based on clothing colors. Our approach differs from standard aesthetic models, since in accounts for a personal variation. In addition, it can process not only lower-order color pairs, but also groups of several colors.

\end{abstract}

\IEEEpeerreviewmaketitle

\section{Introduction}

Understanding human aesthetic preference is a challenging task that can be highly useful for a number of industries, including design \cite{tokumaru}, marketing, and fashion (e.g., refined user-driven results for visual search engines). Fashion aesthetics involves many aspects, including the color, various styles (e.g., sleeves types), materials, spatial compositions, etc. However, consumers usually judge an item within 90 seconds of viewing and initial assessments are mostly driven by colors (i.e. when a human perceives colors, a rich network of associations gets activated). So, we pay a lot of attention to color aesthetics in particular, for making personalized recommendations for images. But we need to remember that preference for harmonious color stimuli is just one factor underlying aesthetic response. 

Color aesthetics involves studying of visually appealing color combinations hidden in the interior, fashion look or even some piece of art. So that a user sees a composition (of any items) and gets the aesthetic pleasure. Color is generally considered to be one of the most important and distinguishing visual features. Additionally, it is often treated as an aesthetic issue, having a significant impact on product sales, accounting for 85\% of the reason why consumer purchases a product. How? By creating an impression and raising the aesthetic senses, color influences decision-making (buy or not to buy) processes in our brains. 

Humans have different levels of visual sensitivity and different color perception abilities. Differentiating between millions of colors, wherein describing a whole image by only using a few colors is a challenging task. That is why the majority of shopping portals, like Amazon and eBay use TBIR, have limitations like subjectivity, manual image tagging and incompleteness. So, we need a human-consistent way to represent color images. 

By adopting fuzzy set based representations and the necessary calculus for them we can solve the problem of semantic gap between low-level color visual features and high-level concepts. In our previous works \cite{jaciii}, \cite{ijufks}, we discuss how we use fuzzy sets to deal with the uncertainty linked to apparels images for the online shopping coordination. Color channels distributions in our space are expressed with fuzzy membership functions \cite{jaciii}. 

We claim that color theories must be shaped by aesthetic norms, including taste (preference for single colors) and trends (context-aware harmonious palettes). We propose a technique to predict the aesthetic preference for color combinations by introducing the new variables, color harmony and color preference. Aesthetic responses to colors are highly influenced by harmony between colors, since the same color can create different impression when viewed together with different colors. We believe that concepts of \textit{preference} and \textit{harmony} and their relationship can serve as a tool for future investigations in aesthetics across multiple domains.

Difference between preference and harmony is crucial. Aesthetic preference for color combinations is mostly driven by color harmony and there are some common tendencies in defining them. However, there are also small differences in the degree to which people prefer harmony, with correlations in the range from 0.03 to 0.75. \cite{palmer2}, \cite{palmer3}. In the context of e-commerce shopping, there are two main related questions:

\begin{enumerate}
\item How do colors of apparels influence preference and harmony judgements? 
\item How to predict a preference for a look based on colors of apparels?  
\end{enumerate}

 In this paper, we try to answer the above questions by predicting combination preference from components preference and harmony. We use results of experiments of Berkeley Color Project \cite{palmer1},  \cite{palmer2},  \cite{palmer3},  and claim that higher individual preference for distinct colors and higher harmony ratings imply higher preference. 

We extract harmonious pallets from big data sets, test them by conducting a questionnaire and then use comparison algorithms based on fuzzy similarity. Our previous method just gave us true/false response on a harmony of a particular combination. In this work we propose the method to quantify the harmony and add the phenomenon of preference.

\section{Motivation}

In our previous works \cite{jaciii}, \cite{ijufks} we proposed an approach based on fuzzy sets and logic towards the creation of a perceptual color space (FHSI, fuzzy HSI). FHSI colors are modelled by means of fuzzy sets defined on an HSI color space and a fuzzy partition is defined in the corresponding color feature domain (fuzzy color space). There are 92 colors in FHSI. The soft boundaries between the color categories  were derived experimentally through an online survey based on human color categorization. We also defined methods for finding the perceptual difference between colors and the degree of similarity between images based on FHSI system \cite{jaciii}, \cite{ijufks}.

Currently we are working on developing a human-consistent image retrieval system for apparel coordination based on the FHSI perceptual color space. We faced the problem of quantitative evaluation of a color combination harmony. 

Humans often experience colors not in an isolation, but in a combination. The aesthetic perception of a color group is strongly influenced by its overall harmony. Hence, it is essential to consider the congruency of chromatic compositions, rather than how much people like single colors\cite{palmer2}.

Throughout history, it has been the contradiction in research works of color theorists studying harmony. In \cite{chevreul} Chevreul  defined the law of simultaneous contrast of colors  and proposed harmonies of analogous and contrasting colors. Another great color theorist, Itten, stated that ``harmonious" color combinations are composed of closely similar chromos (e.g. tones, tints and shades), or else of different colors of the same nuance\cite{itten}. In essence, his theory defines harmonious colors as colors producing neutral gray when mixed. Furthermore, Munsell and Ostwald  put ahead the idea that colors are harmonious if they have some relation in the given color space (e.g., if colors are similar in hue) \cite{palmer2}.

As we see, the color literature is full of discrepancies. If we unite these theories into one system we will obtain that nearly every color combination can be considered as harmonious \cite{palmer2}. The source of this contradiction lies in the fact that there is still neither human consistent color space nor single best representation of color. There are multiple spaces that characterize the color features from different perspectives instead. Therefore, it is a very challenging task to represent, measure and process colors and their harmony.

\section{Evaluation of Color Harmony and Preference}
Preference and harmony are often used interchangeably. However, these two phenomenon are similar only  in case an observer likes the colors in the  combination. To avoid a confusion we need to accurately define and measure them.

\subsection{Typology of Color Judgements}

In order to better customize the system for each user, we need to define and carefully understand the difference between \textit{Single Color Preference (SCP)}, \textit{Color Scheme Harmony (CSH)},  and \textit{Color scheme preference (CSP)}. They are all three distinct types of judgments and different ways of evaluation of perceptual responses to color schemes.

\subsubsection{Single Color Preference}

SCP reflects the contextless preference ratings. Usually, people tend to prefer some colors over others, and nearly everyone has its favourite color. 

\begin{figure}[!t]
\centering
\includegraphics[width=1.6in]{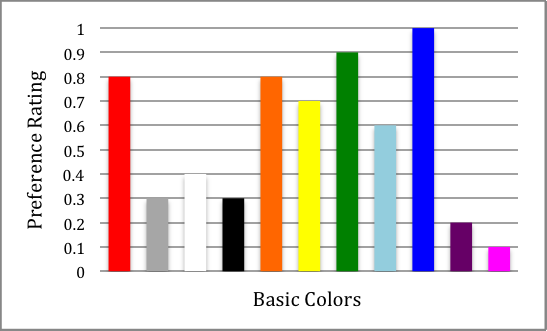}

\caption{Example of Single Color Ratings.}
\label{fig_single_cr}
\end{figure}

SCP measures can be obtained during or after the registration in the system.  We just offer a small survey with simple visual content (colors) with questions (e.g., How much do you like the display?) and collect the responses using a rating scale (e.g., Not at all , Good, Very much, etc.). An example of Single Color Ratings can be seen in Fig. \ref{fig_single_cr}. The idea is to account for basic colors without a considerable loss in precision. 

\subsubsection{Color Scheme Harmony}

In contrast, a CSH reflects how strongly the colors in the combination are going well together, regardless of whether an individual  likes the given combination or not. In other words, CSH indicates the harmony of the color combination as a whole.

\subsubsection{Color Scheme Preference}

A color palette (combination, scheme, group) preference is defined as how much an individual likes a given combination of colors. 
So,  it reflects an aesthetic preference for a given palette as a whole

SCP ratings are not enough to predict CSP. To better define palette preference we need a relational factor like harmony. The problem is how to derive it? Conventional methods are no longer valid to meet current requirements \cite{ijufks}.

\subsection{Harmonious Palettes Derivation }

 In \cite{palmer1},  \cite{palmer2}, \cite{palmer3} it was concluded that color harmony is a function of color similarity. What is more, people tend to prefer color combinations, which contain colors that are similar in hue, cool, and desaturated. However, we believe that harmony is a very complex phenomenon and it is nearly impossible to purely define it in terms of strict rules.
 
We want to extract harmonious, fuzzy color-based pallets from data set. This is done by forming groups containing looks with similar color compositions. To achieve this, we use the fuzzy model we developed, formulas for fuzzy color difference, palettes similarity (described in \cite{jaciii}, \cite{ijufks}).

As a data set we took 10000 images with fashion looks from various sources, including the most popular fashion sites , like polyvore.com, lookbook.nu, instyle.com, dailylook.com and various style communities in SNS (instagram, VK, Facebook). The preference was given to looks having more likes.        

For each image M in a data set we perform the following:

\begin{enumerate}
\item	Compute the fuzzy dominant color histogram of M
\item	Compute the mean average perceptual difference $Dp_{avg}$ between $CH$ and members of each harmonious groups. 
\item	If minimal $Dp_{avg}$  is more than difference threshold, form a new group and add M to it. Otherwise, add M to a group with which M has minimal $Dp_{avg.}$
\item Choose groups having at least 100 similar looks falling into a similar fuzzy color scheme.
\end{enumerate}

Processing of data set took almost 7 hours (1-2 seconds for each image, depending on resolution, current number of groups, etc.). On 8\textsuperscript{nd} thousand the convergence was achieved, since almost all new coming images were falling into existing groups and very few new palettes were added during the processing of the last 2000 images. 

\begin{figure}[!t]
\centering
\includegraphics[width=3.6in]{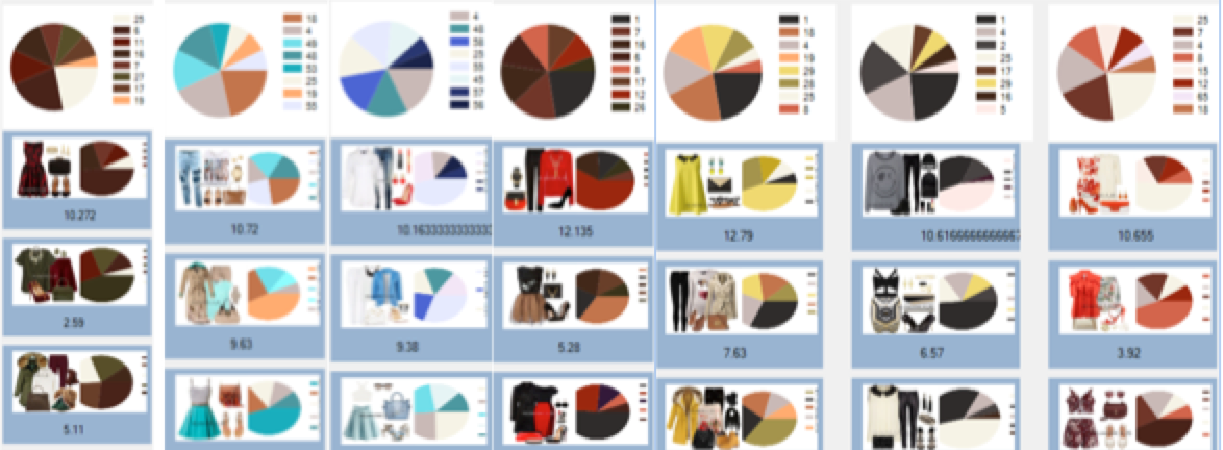}

\caption{Examples of Derived Harmonious Palettes.}
\label{palettes}
\end{figure}

As a result, we got 139 groups in total, 59 of them contained more than 100 images. For each group we took averaged harmonious fuzzy color palette.   Some of the schemes we extracted were similar to Analogous, Contrasting, Triadic, etc. (classified by Itten \cite{itten}), but there were also schemes which are out of any rules.           Examples of harmonious groups with more than 100 similar images can be seen in Fig. \ref{palettes}.

It needs to be emphasized that color preferences may change depending on semantic context \cite{palmer1}. Therefore, derived palettes provide context-aware harmony. We can perform palettes derivation for other domains - art or interior design, for example.  The very essence of this generation process is that it is becoming possible due to FHSI space.

\subsection{User Preference Prediction in Fashion Industry}
How to predict a preference for a look based on colors of apparels?  As mentioned before, a preference for color schemes is influenced by preferences for the component basic colors and ratings of color harmony \cite{palmer1}. Let's combine preferences for the single component colors and ratings of color harmony:

\footnotesize 
\begin{equation}
    Pref(A,B) = \frac{Pref(A)\times w_A+Pref(B)\times w_B}{w_A+w_B} + Harm(A,B) 
     \label{bigeq}
\end{equation}
\normalsize

In Eq. \eqref{bigeq} $w$ is a weight importance of an apparel, $Pref(A), Pref(B)$ are user preferences for single colors $A$ and $B$.  Eq. \eqref{bigeq} also works for three, four, five colors as well.

Generally, aesthetic judgements of fashion looks are influenced by dominance order. For example, dress or skirts usually make more impact on an overall impression rather than accessories. That is why we need to take into account $w$, apparel weight parameter:

\begin{itemize} 

\item For dresses/costumes $w=1$.
\item For up \& down clothes (e.g.,skirts, blouses.) $w=0.75$.
\item For shoes and bags $w=0.5$.
\item For accessories (e.g., glasses, watches) $w=0.25$.

\end{itemize}

Now we can find $Harm(A,B)$ between two fuzzy colors $A$ and $B$. If there is a palette containing both $A$ and $B$, then $Harm(A,B) = 1$. Otherwise we take the most similar color harmonious palette and find the similarity value (which will serve as a harmony value too in this case). Harmony of a group of fuzzy colors which is not in a knowledge base is equal to its similarity with the closest harmonious group. We use similarity measure defined in \cite{jaciii}.

\section{Results}

\subsection{Application} 
The proposed model results in useful predictions of perceived harmony and preference of multi-colored images. 

In a popular application Polyvore, users create looks by dragging items from the menu. It is clear that the color of the new object you add needs to harmonize well with colors of other existing objects, but it is very difficult for a majority of people to choose an object considering such things. 

Using FHSI model \cite{jaciii}, \cite{ijufks} we can index the apparels database. Furthermore, the preference formula Eq. \eqref{bigeq} allows us to sort all the commodity images according to its harmony to the given apparel or look, i.e., the more  harmonious items (suiting the whole look)  will be ranked more ahead. Some of the features include:
   
\begin{figure}[!t]
\centering
\includegraphics[width=3.6in]{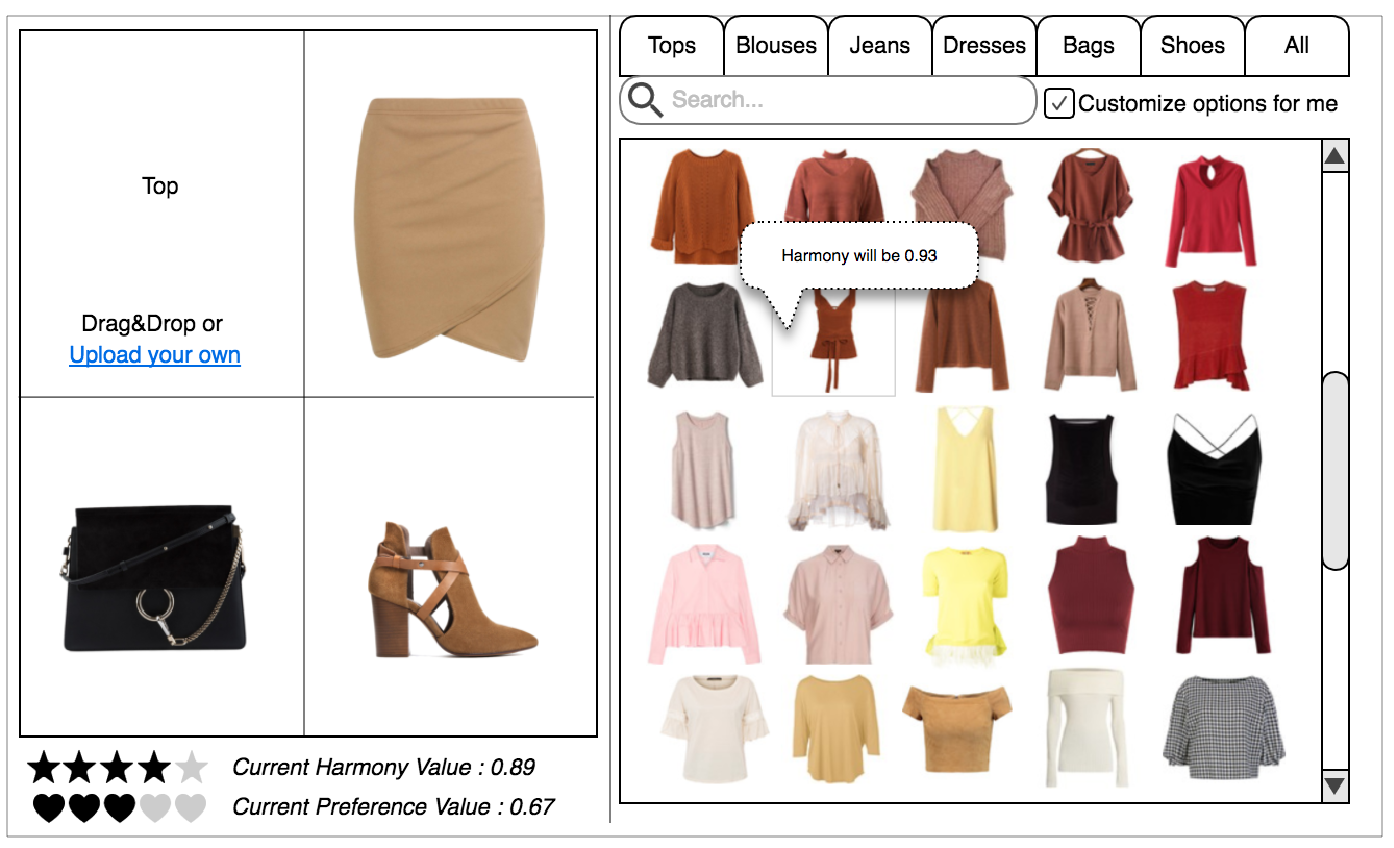}

\caption{Shopping Advisor Application Mockup.}
\label{dd_mockup}
\end{figure}

\begin{itemize}
  \item Having measures for harmony and preference, we can sort the apparels on the right by harmony to the apparels on the left. The suggested apparels  will be unique for each user, since single color ratings allow us to refine color harmony to individual color scheme preference.
  \item Some of the pallets have style names like modern, classic, retro, romantic, elegant, formal, etc. User can choose several styles and see the corresponding apparels only.   
  \item User can input several favourite colors and the system selects colors which are in a harmony and we get many finished looks by the system and also use these harmonious colors to filter the apparels in the menu.

\end{itemize}

The method can also be applied to automatic labelling in large image databases \cite{ch_ii} and customized services. One example for this is furniture coordination (e.g., a user uploads a room image and requests sofas fitting perfectly the interior).

\subsection{Preliminary Experiments}

\textbf{Example 1}. Predict the preference of user $X$ for the look in Fig. \ref{ex1}. As we defined, $w_{dress}= 1$ , $w_{bag}= 0.5$

Since the apparels were preprocessed, we know the ids of fuzzy dominant colors of the dress (A, 12) and the bag (B, 1). Next, suppose the single color ratings of user $X$ areas  represented in Fig. \ref{fig_single_cr}. So, $Pref(A) = 0.8$ and $Pref(B) = 0.5$. $Harm(A, B) = 1$, since fuzzy colors $1$ and $12$ are both in a color harmony palette 27 , see Fig. \ref{ex1}. Finally, according to our preference formula Eq. \eqref{bigeq}, $Pref(A, B)$ for User  $X$ is $0.83$ (value is normalized to a scale [0; 1] ). Note that for an unregistered user (e.g. for a guest)  $Pref(A,B) = Harm(A, B) = 1$.

\begin{figure}
\centering
\begin{minipage}{.25\textwidth}
  \centering
  \includegraphics[width=.7\linewidth]{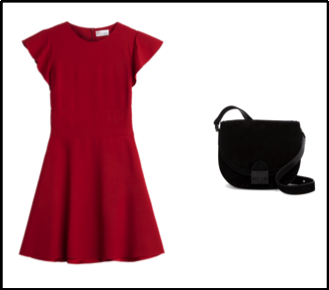}
  \captionof{figure}{Look for Example 1.}
  \label{ex1}
\end{minipage}%
\begin{minipage}{.25\textwidth}
  \centering
  \includegraphics[width=.8\linewidth]{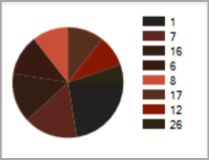}
  \captionof{figure}{Harmonious Palette 27.}
  \label{ex1_palette}
\end{minipage}
\end{figure}

\begin{figure}
\centering
\begin{minipage}{.25\textwidth}
  \centering
  \includegraphics[width=.9\linewidth]{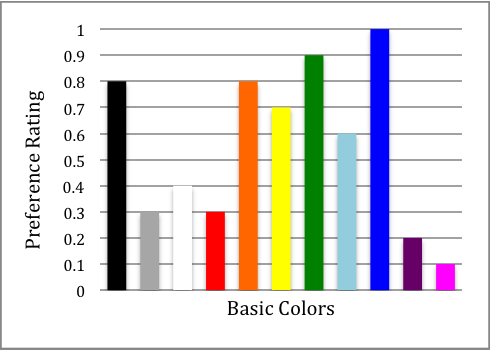}
  \captionof{figure}{Single Color Ratings of User $Y$.}
  \label{single_cr2}
\end{minipage}%
\begin{minipage}{.25\textwidth}
  \centering
  \includegraphics[width=.8\linewidth]{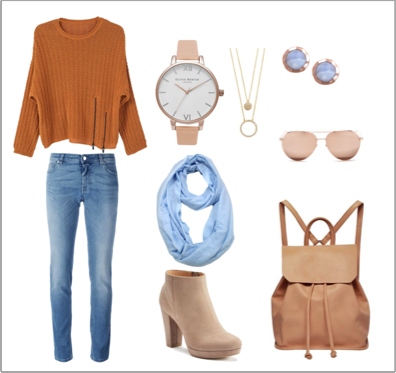}
  \captionof{figure}{Look for Example 2.}
  \label{ex2}
\end{minipage}
\end{figure}

\textbf{Example 2}. Predict the preference of user $Y$ for the look in Fig. \ref{ex2}.
Apparels weights for this example: $w_{up}= w_{down} = 0.75$,  $w_{bag}= w_{shoes} = 0.5$, and $w_{acc} = 0.25$.

Suppose Eq. \eqref{single_cr2} depicts single color ratings of user $Y$. There is no such color harmony group containing  all the fuzzy colors in the look, but there is a group $14$ which is the most similar, similarity is equal to 83\%, so as harmony. According to Eq. \eqref{bigeq}, $Pref(A, B)$ for User  $Y$  normalized to a scale $[0; 1] \approx 0.8$.

%
%
%
%
%
%

\begin{figure}
\centering
\begin{minipage}{.3\textwidth}
  \centering
  \includegraphics[width=.9\linewidth]{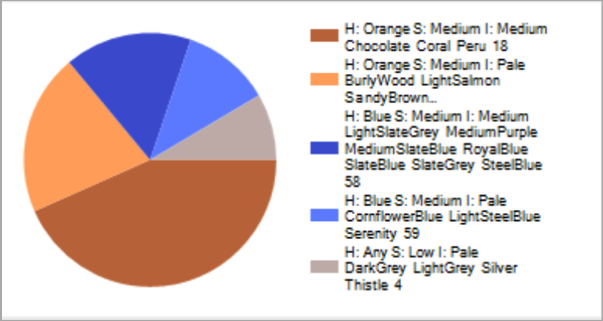}
  \captionof{figure}{Fuzzy Dominant Color \\ Descriptor for Example 2. }
  \label{ex1}
\end{minipage}%
\begin{minipage}{.2\textwidth}
  \centering
  \includegraphics[width=.9\linewidth]{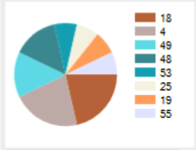}
  \captionof{figure}{Harmonious Palette 14.}
  \label{ex1_palette}
\end{minipage}
\end{figure}

\section{Conclusions and Future Work}
Current work presents an overview of the findings in current research focusing on quantification of the harmony and preference phenomena. The intended application is apparel online shopping coordination system.

The research shown in this paper describes our preliminary results and there are a number of next steps to take in the future. First, we plan to analyse the correlation between preference and harmony ratings through regression analysis. Specifically, the model can be validated by an experiment, where human observers are judging fashion images: whether they are harmonious or disharmonious, and whether they are liked or disliked. We can provide various samples of stimuli and explain the pattern of variation in scheme preferences. Second, we plan to perform the system evaluation by measuring the precision and recall for $\sim$20 queries (Eq. \eqref{p_r}).

\footnotesize
\begin{equation}
    P_{T} = \dfrac{\sum_{i=1}^{T}P_{i} }{T} ,    R_{T} = \dfrac{\sum_{i=1}^{T}R_{i} }{T} , Relevance = \dfrac{P_T}{R_T}
     \label{p_r}
\end{equation}
\normalsize

$T$ is the number of selected queries, $P_i$ = the number of relevant apparels retrieved divided by the number of apparels retrieved,  $R_i$ = the number of relevant apparels retrieved divided by the number of relevant apparels in DB.

Next, for each user we can compute the average difference score as the absolute value of the difference between his real preference ratings and predicted preference. Note that $T$ is a number of looks offered to a survey participant.
\footnotesize

\begin{equation}
     D = \dfrac{\sum_{i=1}^{T} | Pref_{real} - Pref_{pred.}|}{T}
     \label{rel}
\end{equation}
\normalsize

Finally, knowing how to measure preference and harmony properly, we are particularly interested in how much such aesthetic preferences might covary across different semantic contexts. We need to check the palettes relevance in the different aesthetic domains in order to find out whether preferences for harmony are correlated across various domains or not. This will also be the subject of future work.

\end{document}